\documentclass[journal]{IEEEtran}

\usepackage{float} \restylefloat{figure} 
\usepackage{color} 

\usepackage{graphicx}
\graphicspath{{eps/}}
\usepackage{multirow}
\usepackage{colortbl}

\usepackage{fixltx2e}
\usepackage{epstopdf}
\usepackage{rotating}
\usepackage{lscape}
\usepackage{bm} 
\usepackage{cite}
\usepackage{url}
\usepackage[small]{caption}
\usepackage{subfig} 
\usepackage{array} 

\usepackage{amsmath}   
\usepackage{amsfonts}
\usepackage{subfloat}
\usepackage{afterpage}

\newcommand{\bea}{\begin{eqnarray}}
\newcommand{\eea}{\end{eqnarray}}
\newcommand{\be}{\begin{equation}}
\newcommand{\ee}{\end{equation}}
\newcommand{\ba}{\small\begin{array}}
\newcommand{\ea}{\end{array}\normalsize}

\newcommand{\re}[1]{(\ref{#1})}

\begin{document}
\title{Learning Spike time codes through Morphological Learning with Binary Synapses}
\author{Subhrajit Roy, \IEEEmembership{Student Member,~IEEE}, Phyo Phyo San, Shaista Hussain, Lee Wang Wei and Arindam Basu\authorrefmark{1}, \IEEEmembership{Member,~IEEE}
\thanks{Financial Support from MOE through grant MOE2012-T2-2-054 is acknowledged.}
\thanks{Subhrajit Roy, Shaista Hussain and Arindam Basu are with the School of Electrical and Electronics Engineering, Nanyang Technological University, Singapore (e-mail: subhrajit.roy@ntu.edu.sg; shaista001@e.ntu.edu.sg; arindam.basu@ntu.edu.sg). Phyo Phyo San is with the Institute for Infocomm Research, Singapore (e-mail: sanpp@i2r.a-star.edu.sg). Lee Wang Wei is with the SINAPSE laboratory at National University of Singapore (e-mail: lee\_wang\_wei@u.nus.edu). \authorrefmark{1}Arindam Basu is the corresponding author.}
}

\maketitle

\begin{abstract}
 In this paper, a neuron with nonlinear dendrites (NNLD) and binary synapses that is able to learn temporal features of spike input patterns is considered. Since binary synapses are considered, learning happens through formation and elimination of connections between the inputs and the dendritic branches to modify the structure or ``morphology" of the NNLD. A morphological learning algorithm inspired by the `Tempotron', i.e., a recently proposed temporal learning algorithm--is presented in this work. Unlike `Tempotron', the proposed learning rule uses a technique to automatically adapt the NNLD threshold during training. Experimental results indicate that our NNLD with 1-bit synapses can obtain similar accuracy as a traditional Tempotron with 4-bit synapses in classifying single spike random latency and pair-wise synchrony patterns. Hence, the proposed method is better suited for robust hardware implementation in the presence of statistical variations. We also present results of applying this rule to real life spike classification problems from the field of tactile sensing.
\end{abstract}

\begin{IEEEkeywords}
spiking neuron, tempotron, binary synapse, dendrites, plasticity, learning
\end{IEEEkeywords}

\section{Introduction}\label{sec.intro}
Though the representation of stimulus by the neurons in our brain is a topic of much ongoing research and debate, it is widely believed that the timing of the action potentials or spikes fired by these neurons carry important information \cite{Gütig2006}.  Spike latency codes i.e. delay in the spike time after stimulus presentation have been suggested for tactile, olfactory and retinal systems \cite{Johansson2004}. They are also thought to offer significant advantages in terms of reducing power needed for communicating spikes as well as allowing rapid processing of inputs. Hence, neuromorphic engineers, who aim to mimic the brain's processing capabilities in silicon, have also been interested in spike timing based neural networks.

Several analog CMOS integrated circuits operating in the sub-threshold regime have been designed in the past to implement somatic and synaptic functions\cite{Basu2008,Arthur2007,Heidelberg1,Bordeaux1}. However, with the increase of statistical variations due to the constantly decreasing feature size of transistors, performance of silicon neural networks requiring accurate setting of a ``weight" parameter become strongly compromised. The typical solution for this problem is to increase transistor size. However, this alone might not be sufficient to guarantee good matching across the chip. For example, \cite{miniDAC_teresa} demonstrates that $5\mu mX5\mu m$ transistor based 5-bit DACs fabricated in $0.35 \mu m$ CMOS exhibit only $1.1$ effective bits. Calibration techniques can be used to improve the accuracy; however, this incurs significant area penalty due to storage of calibration bits and is unacceptable in large scale systems. The problem of mismatch is also exacerbated for several nanometer scale non-CMOS devices (e.g. memristor\cite{memristor1,memristor2,memristor3} or domain wall magnets (DWM)\cite{dwm_kaushik}) that have potential for use in large scale neuromorphic applications. For example, DWM synapses are expected to have typical programming accuracies of 4-bits which can reduce to 2-3 bits effective resolution due to mismatch \cite{dwm_kaushik}.  Hence, there is a strong need to develop algorithms and architectures that retain the performance of earlier spiking systems but require low-resolution weights.

In this paper, a hardware friendly morphological learning rule for neurons with nonlinear dendrites (NNLD) and binary synapses for classifying spatiotemporal spike patterns is presented. 
Previous studies have shown that NNLD can be successfully applied to learn both mean rate encoded inputs \cite{Poirazi2001,Hussain2013} and spike-timing information \cite{Roy2013,LSM-DER}. Although in \cite{Roy2013,LSM-DER} we proposed a spike time based learning rule for training a NNLD architecture, however the rule was not optimized for memory capacity. Here we propose a novel memory capacity optimized spike timing based learning rule for training NNLD. Our work is inspired by the Tempotron learning rule for spiking neurons \cite{Gütig2006}. However, unlike the Tempotron learning rule that requires weights with high resolution, the proposed network uses low-resolution non-negative integer weights and learns through modifying connections of inputs to dendritic branches. Thus the `morphology' or structure of the neuron (in terms of connectivity pattern) reflects the learning. This results in easier hardware implementation since a low-resolution non-negative integer weight of $W$ can be implemented by activating a shared binary synapse $W$ times through time multiplexing schemes like address event representation (AER)\cite{Boahen2000,Brink2013}. Furthermore, the spiking threshold of the neuron employed in Tempotron is fixed throughout the learning. On the other hand, the proposed method is equipped with a threshold adaptation mechanism trying to optimize the number of false positives and false negatives. Some initial results of this concept were presented in \cite{Phyo2013}. In this paper, we present a novel threshold adaptation technique, more detailed set of results, comparisons with other work and application to a real world problem.

The organization of this paper is as follows: in Section \ref{sec.NLD} and \ref{sec.learning.algorithm}, the architecture of the neuron with nonlinear dendrites and its morphological learning algorithm are introduced. Two spike time based binary classification tasks are discussed in Section \ref{sec.morphological.classify}A and performance of our method in comparison to Tempotron is shown in Section \ref{sec.morphological.classify}B, \ref{sec.Results.comp.Tempotron} and \ref{sec.application.example}. Finally, our work is compared with other classifiers in Section \ref{sec.discussion} and then a conclusion is drawn in Section \ref{sec.Conclusion}.

\section{Neuron with Nonlinear Dendrites}\label{sec.NLD}
\begin{figure}[t] 
	\centering
	\includegraphics[width=0.48\textwidth,height=4cm]{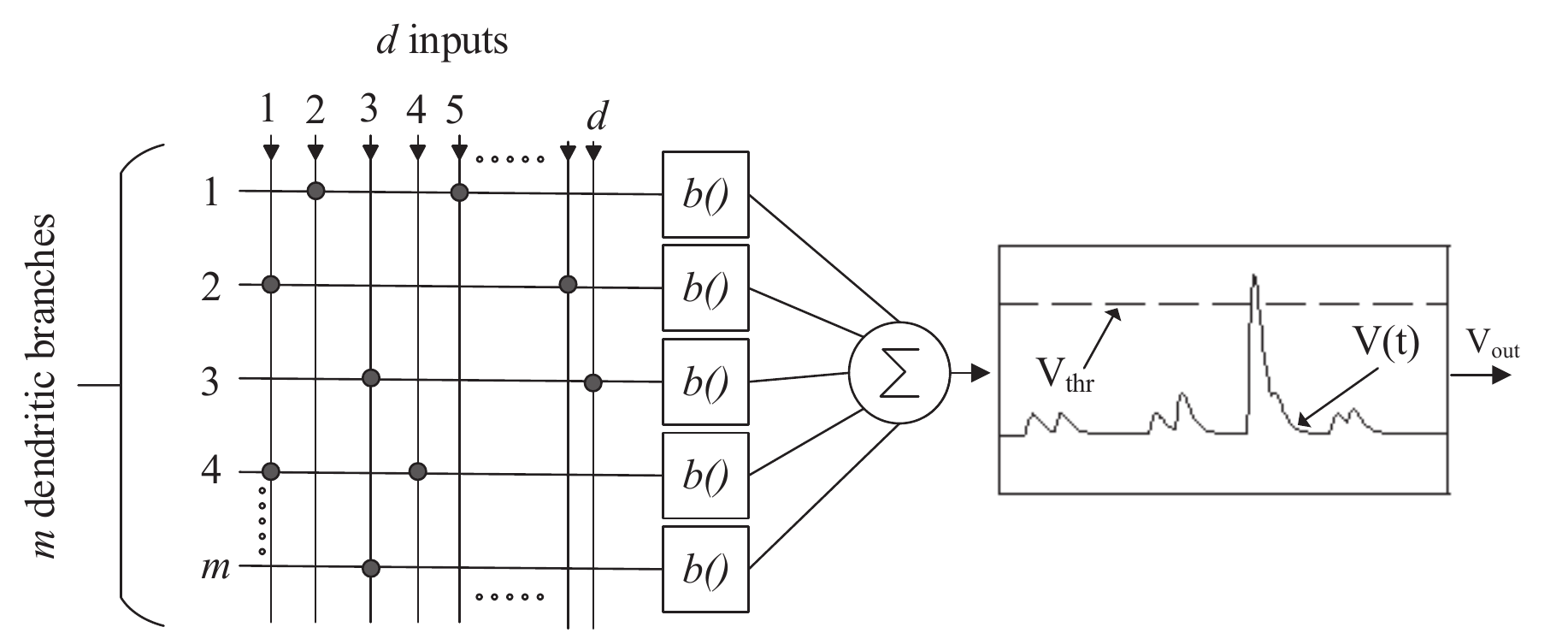}
	\caption{The architecture of neuron with nonlinear dendrites (NNLD).} 
	\label{fig.dendrities}
\end{figure}
It has been shown that a neuron with lumped dendritic nonlinearities possesses higher storage capacity than its counterparts with linear dendritic summation \cite{Poirazi2001}. As presented in Fig. \ref{fig.dendrities}, the structure of NNLD is characterized by $m$ identical dendritic branches and $k$ excitatory synaptic contacts per branch. For each branch, the synaptic contact is formed by one of $d$ dimensions of input afferents where $d>>k$. At the relevant times governed by incoming spikes, the synapses are activated and the membrane voltage is calculated by weighted sum of postsynaptic potentials (PSPs) as follows:
    \begin{equation}\label{eq.Vt}
        V\left( t \right) = \sum\limits_{j = 1}^m {b(v_j(t))}= \sum\limits_{j = 1}^m {b\left( {\sum\limits_{i = 1}^d {{w_{ij}}\sum\limits_{f} K\left( {t - t_{f}^i} \right)} } \right)}
    \end{equation}
where $w_{ij}$ is the weight of the $i^{th}$ synapse formed on the $j^{th}$ branch, $v_j(t)$ is the input to the $j^{th}$ dendritic nonlinearity, $b\left(  \cdot  \right)$ is the nonlinear activation function of the dendritic branch which is characterized by $b\left( v_j(t) \right) = {\raise0.7ex\hbox{${{v_j(t)^2}}$} \!\mathord{\left/{\vphantom {{{v_j(t)^2}} {{x_{thr}}}}}\right.\kern-\nulldelimiterspace}\!\lower0.7ex\hbox{${{x_{thr}}}$}}$ \cite{Hussain2013}, $K$ denotes the post-synaptic potential kernel and $t_{f}^i$ are times of incoming spikes on the $i^{th}$ afferent. Since we consider binary synapses, $w_{ij}\epsilon\{0,1\}$. To suppress unrealistically large values, we include a saturation level $x_{sat}$ at the output of each dendrite such that for $b(v_j(t))>x_{sat}, b(v_j(t))=x_{sat}$. Similar to our earlier work\cite{Hussain2013}, we allow each input afferent to make multiple synaptic connections on the same dendritic branch but limit the total number of connections per branch by enforcing $\sum_{i=1}^{d}w_{ij}=k$ for each $j$. In this work, we consider normalized PSP of the form:
    \begin{equation}\label{eq.K}
    K\left( {t - {t_f}} \right) = {V_0}(\exp [{{ - (t - {t_f})} \mathord{\left/    {\vphantom {{ - (t - {t_f})} \tau }} \right.\kern-\nulldelimiterspace} \tau }] - \exp [{{ - (t - {t_f})} \mathord{\left/    {\vphantom {{ - (t - {t_f})} {{\tau _s}}}} \right.    \kern-\nulldelimiterspace} {{\tau _s}}}])
    \end{equation}
where the parameters $\tau$ and ${\tau _s}$ =${\tau  \mathord{\left/ {\vphantom {\tau  4}} \right. \kern-\nulldelimiterspace} 4}$ denote the decay time constants of membrane integration and synaptic current respectively\cite{Gütig2006}.

For binary classification, the final output voltage $V_{out}$ is interpreted to take a value of either ``1'' or ``0'' depending on whether the summed membrane voltage $V(t)$ is crossing a threshold voltage $(V_{thr})$. Similar to \cite{Gütig2006}, this indicates that a neuron fires at least one spike if $V(t)$ crosses $(V_{thr})$; otherwise it remains quiescent. After the neuron fires a spike, it is returned to the refractory state and the rest of the spikes in the incoming pattern do not affect the computation.

\section{Morphological Learning Algorithm}\label{sec.learning.algorithm}
On one hand the proposed learning rule tries to optimize the connections between the input lines and the synapses while on the other hand it tries to find a optimal value of the neuron firing threshold$(V_{thr})$. Below we describe both the processes.
\paragraph{Learning the connections}
Since we have binary synapses, a morphological learning rule that can modify the connections between afferent lines and synapses is needed. The inspiration of this work comes from the Tempotron learning rule \cite{Gütig2006} that learnt the temporal feature of random spike patterns in two classes by updating its weights so that the neuron fires a spike for patterns in class $P^+$ and stays silent for the other class. It was shown in \cite{Gütig2006} that the Tempotron rule endows a neuron with larger classification capacity than a perceptron with same number of synapses. Hence, we also start with a cost function that measures the deviation between the maximum membrane voltage ($V_{max}$) generated by misclassified patterns and $V_{thr}$ defined as:
\begin{equation}\label{eq.error.function}
    E = \left\{ {\begin{array}{*{20}{c}}
    {{V_{thr}} - {V_{{{max }}}},\quad if\;pattern\;in\;{P^ + }\;is\;presented}\\
    {V_{{{max }}}} - {{V_{thr}} ,\quad if\;pattern\;in\;{P^ - }\;is\;presented}
    \end{array}} \right.
    \end{equation}
where $V_{max}$ is the maximal value of postsynaptic potential $V(t)$ at the time $t_{max}$, i.e., ${V_{\max }} = V\left( {{t_{\max }}} \right)$. According to the gradient-descent method, the change in synaptic efficacy for the first case is calculated by:
\begin{equation}\label{rule}
\begin{aligned}
\Delta {w_{ij}} = & -\frac{\partial {E}}{\partial {w_{ij}}}\\
                = & -\frac{{\partial \left( V_{thr}-\left( {\sum\limits_{j = 1}^m {b\left( {\sum\limits_{i = 1}^d {{w_{ij}}\sum\limits_{f} K\left( {t_{max} - t_{f}^i} \right)} } \right)} } \right) \right) }}{{\partial {w_{ij}}}}\\
                = &\; b^{'}{(v_j(t_{max}))} \sum\limits_{f} K (t_{max} - t_{f}^i) 
\end{aligned}               
\end{equation}
where $b'(.)$ denotes the derivative of $b(.)$. The gradient for second case can be calculated similarly and we do not show it here for brevity. As mentioned earlier, since we consider binary weights, i.e., $w_{ij}=1$ if a connection exists and $0$ otherwise, we cannot directly modify the weights by adding the $\Delta w_{ij}$ term derived here. Instead, the term $\Delta {w_{ij}}$ in Equation \ref{rule} is reinterpreted as a correlation term $c_{ij}(=w_{ij})$
and is used to guide the process of swapping connections. At every iteration of the learning process, the synapse with the lowest $c_{ij}$ averaged over an entire batch of patterns from a randomly chosen target set will be replaced (weight changed from 1 to 0) with the highest $c _{ij}$ synapse in a candidate replacement set similar to \cite{Hussain2013}. To keep the paper self-contained, the mechanism of the learning process is outlined briefly below:
\begin{enumerate}
    \item The learning process starts with random initialization of the connection matrix between input afferents and dendritic branches.
    \item In each iteration of the training process, the activation of synapses on each dendritic branch are determined and the cell membrane output voltage ($V(t)$) in \re{eq.Vt} is calculated for all the input patterns.
    \item From the calculated $V(t)$, the maximum membrane voltage ($V_{max}$) is observed and classification result is determined, i.e., the patterns are correctly classified if ${V_{\max }} > {V_{thr}}$ for $P^+$ and ${V_{\max }} < {V_{thr}}$ for $P^-$.
    \item A random set $T$ of $n_T$ synapses having weight 1 was targeted for possible replacement. For all misclassified patterns, the correlation term, $c _{ij}$ is calculated for each synapse in $T$ and averaged over the entire pattern set, 
    \item The poorest-performing synapse (minimum $c _{ij}$) in $T$, $T_{min}$ is chosen for replacement. 
    \item To aid the replacement process, a randomly chosen set $R$ containing $n_R$ of the $d$ afferent lines is forced to make silent synapses with weight 1 on the dendritic branch of $T_{min}$. These synapses are ``silent" since they do not contribute PSP to the computation of $V(t)$--so they do not alter the classification when the same pattern set is re-applied. But now $c_{ij}$ is calculated for synapses in $R$ and $T_{min}$ is replaced with the best-performing synapse (maximum $c _{ij}$) in $R$. 
    \item The learning from step (2) to (6) continues until all the patterns are correctly learnt (or) the maximum number of iteration is reached.
\end{enumerate}

\paragraph{Learning the threshold}
Since we do not have an arbitrary multiplicative weight in our neural model, the range of maximum voltages obtainable from our model in response to a fixed temporal spike pattern is limited. This is similar to the problem faced in \cite{Hussain2012}. Hence, improper selection of threshold may largely degrade the classification performance since a very large $V_{thr}$ ($= m\times k\times K_{max}$ for example) may never be crossed by $V(t)$. In \cite{Phyo2013}, we determined the value of $V_{thr}$ by noting the maximum value of $V(t)$, i.e., $V_{max}$ at time $t_{max}$ for a large number of random input spike patterns and connection matrices. The resultant probability distribution of $V_{max}$ was used to determine the optimal threshold $V_{thr}$. In \cite{Phyo2013}, $V_{thr}$ was set to be the voltage corresponding to the peak of probability distribution function. Here, we propose an automatic mechanism for adapting $V_{thr}$ during training. This technique involves updating the value of $V_{thr}$ after each iteration which is guided by the following formula:
\begin{equation}\label{thr_adapt}
\Delta V_{thr}= \eta (w_{fp}FP-w_{fn}FN)
\end{equation}	
where $FP$, $FN$, $w_{fp}$, $w_{fn}$ and $\eta >0$ are the number of false positives, number of false negatives, weightage associated with false positive error, weightage associated with false negative error and threshold learning rate respectively. In this article, we keep $w_{fp}=w_{fn}=1$. Equation \ref{thr_adapt} is responsible for balancing the number of false positives and false negatives. When $FP>FN$, the number of negative patterns incorrectly classified as positive patterns is more than the number of positive patterns incorrectly classified as negative patterns so to balance $FP$ and $FN$ Equation \ref{thr_adapt} increases the value of $V_{thr}$. On the other hand, when $FP<FN$ Equation \ref{thr_adapt} diminishes the value of $V_{thr}$. The rate of threshold adaptation is controlled by the threshold learning rate $\eta$.      

\section{Experiments and Results}\label{sec.morphological.classify}
\subsection{Problem Description}\label{task_descrip}
In this sub-section, we describe the two tasks used to demonstrate the performance of our algorithm. The reason for this choice is that both of these are standard problems shown in \cite{Gütig2006} and facilitates comparison.  
\subsubsection{Task I: Classifying Random Latency Patterns}\label{sec.classify.random.latency.pattern}
The first task is binary classification of single spike random latency patterns \cite{Gütig2006}. To perform the task, $P$ spike patterns were generated and randomly assigned to one of the two classes ${P^ + }$ (Class 1) or ${P^ - }$ (Class 2). Each spike pattern ${\rm X} = \left( {{x_1},{x_2}, \ldots ,{x_d}} \right)$ consists of $d$ afferents, where each of them spiked only once at a time drawn independently from a uniform distribution between $1$ and $T$ ms. 

\subsubsection{Task II: Classifying pairwise Synchrony Patterns}\label{sec.classify.PairWise.pattern}
To examine the ability of our algorithm to learn correlations in multiple spikes, another data set that consists of pairwise synchrony events in each pattern is generated. In this data set, all the $d$ afferents are grouped into $\left( {{d \mathord{\left/ {\vphantom {d 2}} \right. \kern-\nulldelimiterspace} 2}} \right)$ pairs and afferents in a given pair fire single spike patterns synchronously. Since synchronous events occur at random, uniformly distributed times in both pattern categories, so that class information is embedded solely in the patterns of synchrony; neither spike counts nor spike timing of individual neurons carry any information relevant for the classification task. This task mimicked spike synchrony-based sensory processing. 

For both Task I and II, we have kept $d$ as 500.

\subsection{Results: Performance of NNLD trained by morphological learning algorithm}\label{sec.Results}

Throughout the experiment, the design parameters $m$, $k$, $x_{thr}$, $x_{sat}$, $\tau$ and $T$, were chosen as $100$, $5$, $1$, $100$, $15$ $ms$ and $400$ $ms$  respectively. The detailed criteria for selection of parameters $x_{thr}$ and $x_{sat}$ is described in \cite{Hussain2013}. As discussed in Section \ref{sec.learning.algorithm}, the firing threshold $V_{thr}$ is adapted and so it is randomly initialized before training.

\begin{figure}[t] 
	\centering
	\subfloat[]{\includegraphics[width=0.25\textwidth]{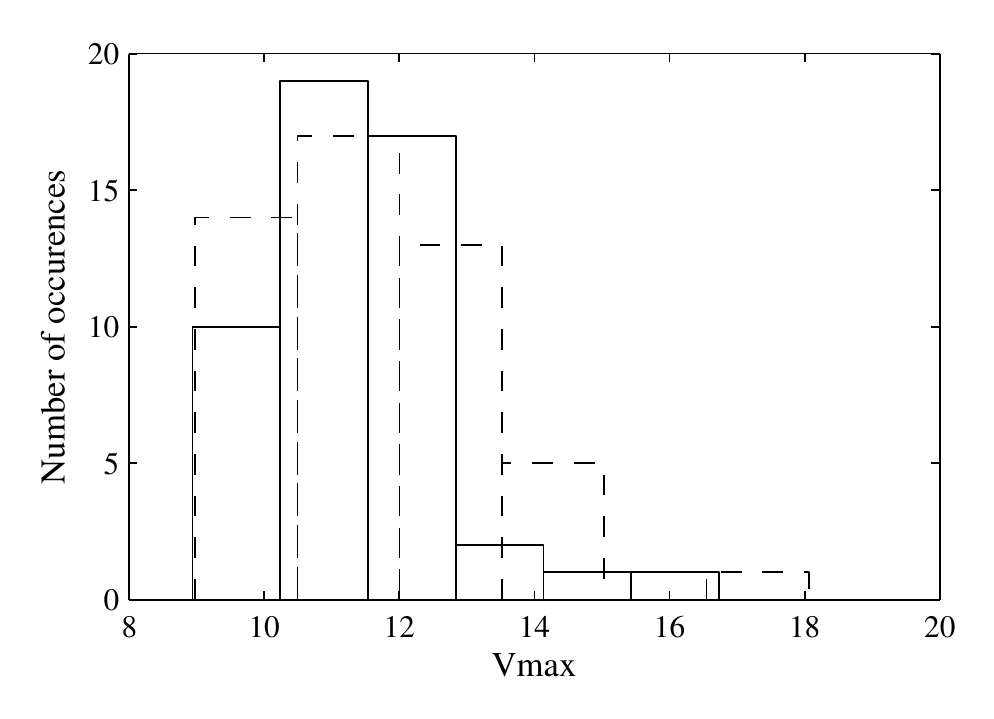}}
	\subfloat[]{\includegraphics[width=0.25\textwidth]{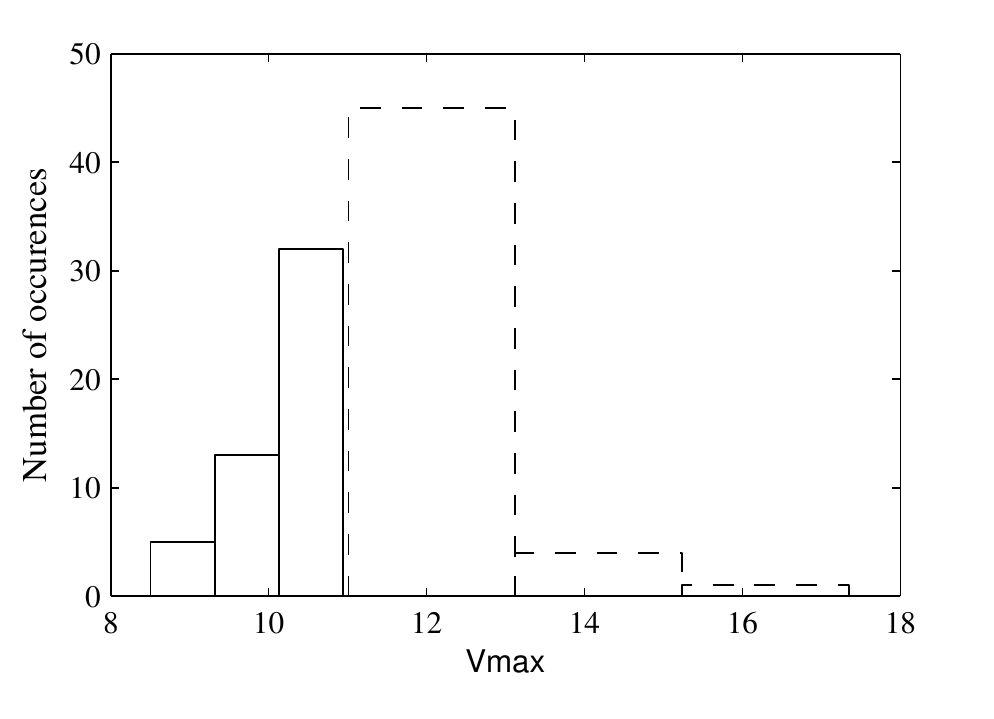}}
	\caption{Distribution of $V_{max}$ for 100 patterns (a)before and (b) after training for Task I. $V_{max}$ for patterns in $P^+$ and $P^-$ are shown in dashed and solid line respectively.} \label{fig.100PVmaxDistribution}
\end{figure}

To start with Task I, the NNLD in Fig. \ref{fig.dendrities} was trained on a small number ($=100$) random latency patterns as generated in Section \ref{sec.classify.random.latency.pattern}. The results in Fig. \ref{fig.100PVmaxDistribution} (a) and (b) show that the proposed method can efficiently perform the classification task. A clear separation between Class 1 and Class 2 shows that the proposed method is able to respond to the random single spike latency patterns by shifting $V_{max}$ closer and farther to the $V_{thr}$ for each pattern in classes 1 and 2 respectively.

\begin{figure}[t]
        \centering 
        \includegraphics[width=0.5\textwidth]{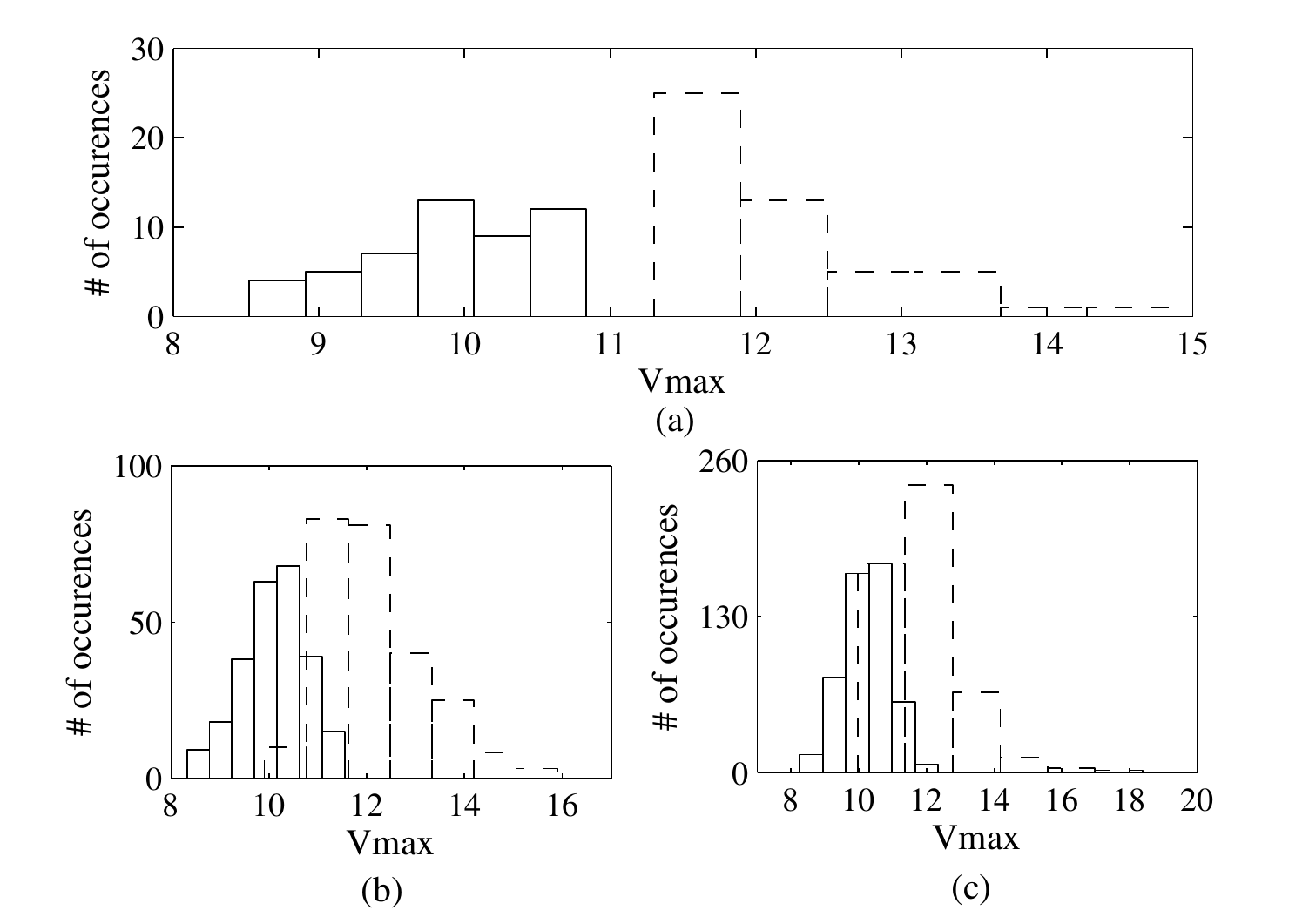}
        \caption{The distribution of $V_{max}$ for a set of (a) 100, (b) 500 and (c) 1000 patterns in task I showing 100\%, 94.8\% and 84.9\% accuracy respectively.} \label{fig.VmaxDis.200to1000P.LP}
    \end{figure}

The NNLD was also trained for larger number of input patterns ($500$ and $1000$ patterns) as presented in Fig. \ref{fig.VmaxDis.200to1000P.LP} (b) and (c). It shows that the proposed method can perform the classification task quite well by achieving accuracies of $95.58 \% (SD=0.54\%)$ and $86.57 \% (SD=0.72\%)$ respectively for these cases.
Next, we performed the experiment of Task II with our network. It was observed that the classification performances (100$\%$, 100$\%$ and 99.71 $\% (SD=0.1\%)$ accuracy for 100, 500 and 1000 patterns) are much better than that of performance in random latency patterns of Task I.Hence, our method can identify the extra information embedded in synchrony of neural firings. It also reveals that the learning rule does not depend only on a single synchronous pair but on a huge number of synchronous pairs.
    

\subsection{Results: Comparison with Tempotron learning}\label{sec.Results.comp.Tempotron}
\begin{figure}[t] 
        \centering
        \includegraphics[width=0.48\textwidth]{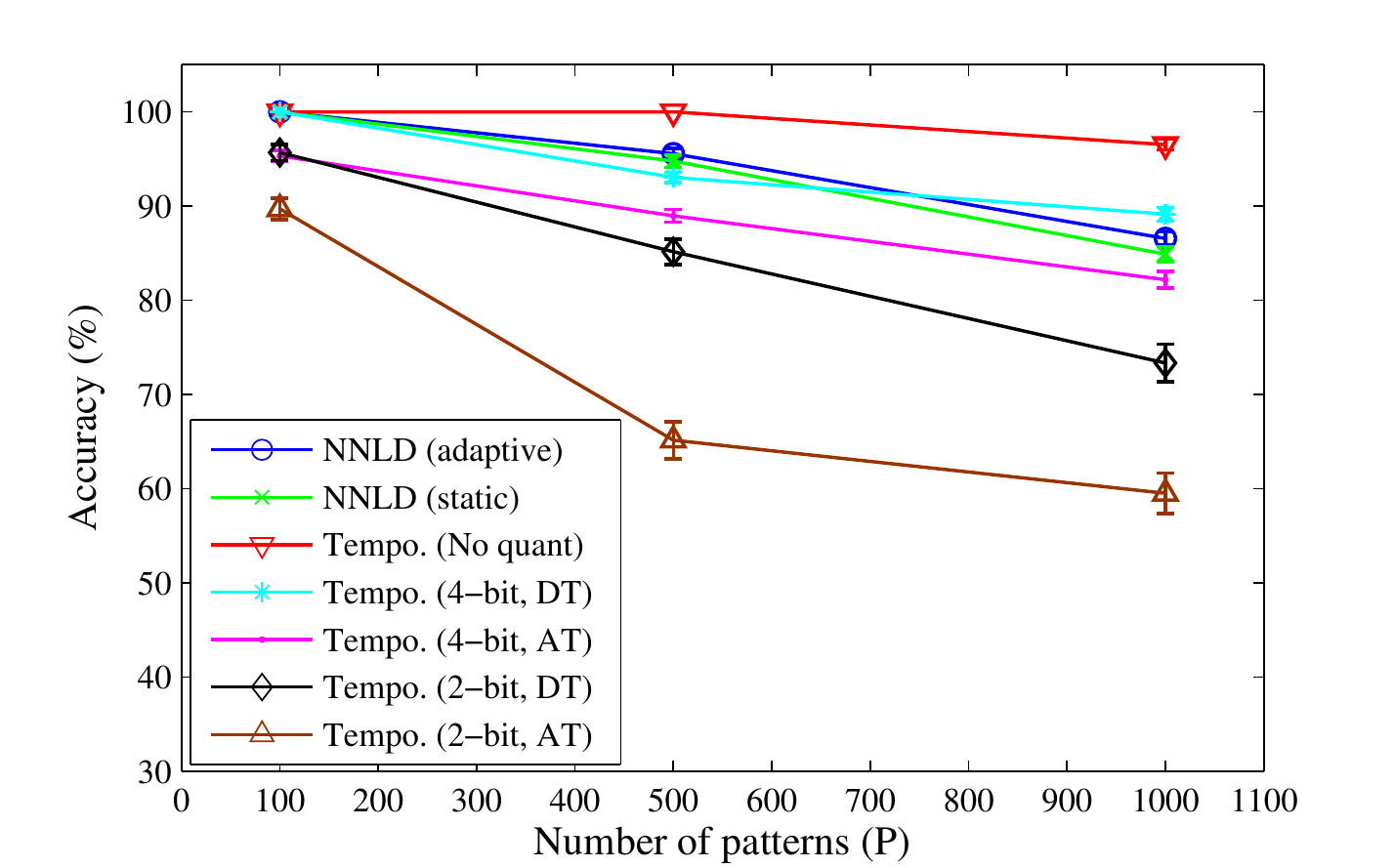}
        \includegraphics[width=0.48\textwidth]{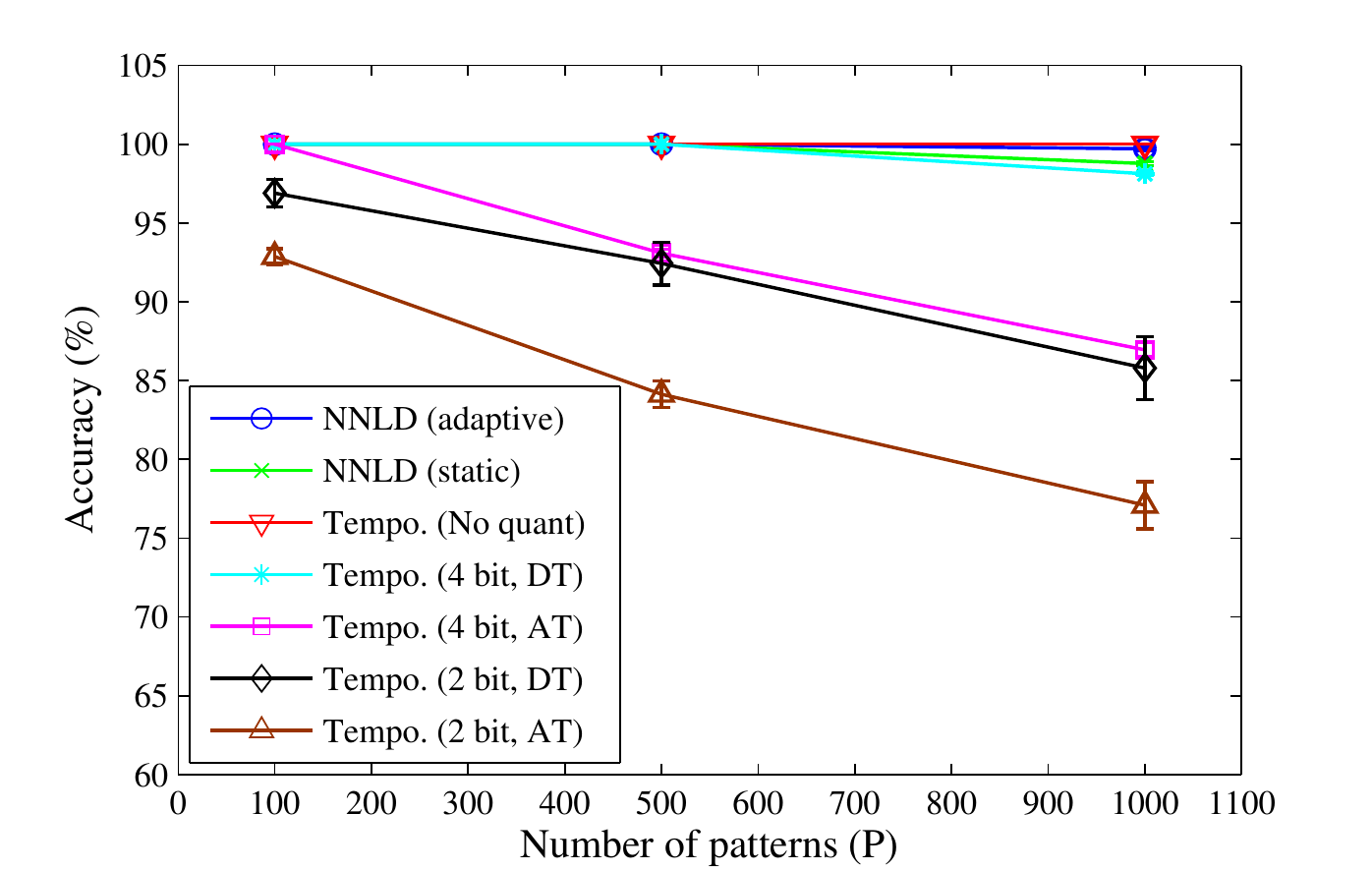}
        \caption{Comparison studies on classification performance, for Tempotron learning \cite{Gütig2006} and the proposed morphological learning rule for (a) Task I of Random Latency Patterns and (b) Task II of Pair-Wise Synchrony Patterns. The results have been averaged over 10 independent trials.} \label{fig.Quantization}
    \end{figure}

Next, the performance of proposed method is compared with the Tempotron learning \cite{Gütig2006} that learns spike time patterns by using weight updates. The number of synapses used by Tempotron is equal to the number of input afferents $d$. Since we are interested in the performance of these algorithms in their hardware implementations plagued by mismatch, we consider the performance of the Tempotron when its weight is quantized at different resolutions. Further, we do the quantization in two ways: either after training (AT) or as a step during training (DT). The first method corresponds to the case where weights trained in software version of the algorithm is downloaded to the hardware while the second method is analogous to performing training on-chip. Furthermore, for comparison with our previous threshold selection technique \cite{Phyo2013}, we have also calculated the value of $V_{thr}$ by equating it to the voltage corresponding to the peak location of $V_{max}$ distribution over $10000$ random input spike patterns. We term this as $V_{thr,static}$. Fig. \ref{fig.Quantization} depicts that the performance obtained by morphological learning with adaptive threshold is superior to that of learning with fixed $V_{thr,static}$. Moreover, the comparison results in Fig. \ref{fig.Quantization} (a) also show that the Tempotron using floating-point numbers achieves better performance compared to the proposed method. However, when the high resolution weights are quantized at 2-bit level, its performance is worse than the proposed method. Also, it can be seen that the performance is better when the quantization is performed within the learning loop. This is to be expected since the learning algorithms can now try to correct this quantization error as well.

Only at 4-bit quantization level, the classification performance of Tempotron ($93.05\%(SD=0.55\%)$ and $89.14\%(SD=0.7\%)$ accuracy for $500$ and $1000$ patterns) in Task I is comparable to our proposed method using 1-bit or binary weights. This underlines the importance of our proposed method in robustly implementing spike timing based classifiers using low-resolution analog synapses available in nano-scale CMOS or non-CMOS devices. Similarly, the classification performance of NNLD for Task II with 1-bit binary weights in Fig. \ref{fig.Quantization} (b) is comparable to performance of Tempotron at 4-bit quantization which shows about $100$ $\%$ and $98.12\%(SD=0.12\%)$ accuracy respectively for $500$ and $1000$ patterns.

\section{Application Example}\label{sec.application.example}
Till now the proposed algorithm has been applied to classify synthetic synchrony patterns. But, to check whether the algorithm can work in real world problems, it is used to classify Tactile information. The patterns which have been till now presented for classification had one spike per afferent but real world scenarios may have multiple spikes per afferent.
\paragraph{Task Description}
A detailed description of the task can be found in \cite{Lee2003}--here, we give a brief description for completeness. The task requires a flexible, stretchable and conformable tactile sensor array made up of conductive fabric which is used for data collection. Two glass spheres (indenters) of diameter $65 $ mm and $105$ mm were indented onto this sensor array by a suitably programmed $6$-axis robotic arm. The indentation force used was $4$N, as measured using a load cell placed below the sensor having an accuracy of $0.01$N. The signal recording was started just before placing the indenter onto the sensor array and stopped before removing the indenter. Between consecutive indentations, a $5$ second pause was provided so that the sensors were able to recover. A total of $100$ recordings were taken per indenter. The collected data were converted to spikes as described in the next sub-section. For each indenter, $60$ randomly chosen recorded data was used for training both the proposed algorithm and Tempotron. The remaining $40$ recordings were used for testing.
\paragraph{Spike Train Generation}\label{spike_gen}
The conversion of the analog data recorded by the sensor array to spike trains involves the following steps. First, the analog data is converted to digital output by applying a fixed threshold of $0.5$. This digital output is inverted to generate another set of digital data. Each channel of the digital data and its inverted version are passed through Leaky Integrate and Fire Neuron to generate two sets of spike trains. These spike trains are combined to form the spike response to be given as an input to the algorithms. The usage of both the digital data and its inverted version for spike generation ensures both low to high and high to low transitions are captured by a change in firing activity. The combined spike train patterns given as input to the algorithms consist of 130 afferents. Thus, we allot 130 synapses for both Tempotron and NNLD.

\paragraph{Results}
The performance of the proposed method on this application is compared with Tempotron algorithm at different quantization levels in Table \ref{table:MCsimulationResultsSquareBlock}. The results, averaged over 10 independent trials, show that the proposed algorithm is able to achieve an accuracy of $96.54\%(SD=0.6543\%)$. Although Tempotron without quantization performs better ($97.06\%(SD=0.5123\%)$) than the proposed learning rule on NNLD, but after quantization, at least $6$ bits of weight resolution is needed by Tempotron to match our performance with $1$ bit weights.


\begin{table}
	\caption{Performance comparison of morphological learning rule on NNLD and Tempotron}
	\centering
	\begin{tabular}{|c|c|c|c|c|}
		\hline
		\multicolumn{1}{|>{\columncolor{white}}c|}{Cases}&\multicolumn{2}{c|}{Accuracy}\\
		\arrayrulecolor{black}
		\arrayrulecolor{black}
		\cline{2-3}
		\multicolumn{1}{|>{\columncolor{white}}c|}{}&Mean & SD\\
		\hline
		Morph. lear. on NNLD (adaptive $V_{thr}$)&96.54& 0.6573\\
		\hline
		Morph. lear. on NNLD (for $V_{thr\_ static}$)&95.64 & 0.6473\\
		\hline
		Tempotron (no quantization) \cite{Gütig2006}   & 97.06& 0.5123   \\ 
		\hline
		Tempotron (6-Bit quantization after training)   & 95.64& 0.7614 \\ 
		\hline
		Tempotron (6-Bit quantization during training)   & 96.26& 0.6822 \\ 
		\hline
		Tempotron (4-Bit quantization after training)   & 78.87& 1.8371 \\ 
		\hline
		Tempotron (4-Bit quantization during training)   & 85.92& 1.3846 \\ 
		\hline
		Tempotron (2-Bit quantization after training)   & 51.56& 3.765 \\ 
		\hline
		Tempotron (2-Bit quantization during training)   & 55.88& 2.8731 \\
		\hline
	\end{tabular}
	\label{table:MCsimulationResultsSquareBlock}
\end{table}

\section{DISCUSSION}\label{sec.discussion}
Here, we compare our method with other reported algorithms and other possible variants of morphological learning.
\subsection{Comparison to other supervised spiking neural classifiers}\label{sec.alternative.cij}
In recent years, several supervised learning algorithms have been proposed such as Tempotron \cite{Gütig2006}, ReSuMe \cite{ReSuMe_neco}, SpikeProp \cite{SpikeProp} and Chronotron \cite{Chronotron} for training spiking neurons. From a pattern recognition viewpoint, these algorithms can be classified into two types. The general theme of the first type of algorithms is that a desired output spike train is specified prior to learning for each class of patterns. SpikeProp, ReSuMe and Chronotron are examples of this type, among which SpikeProp can only produce a single output spike whereas the others are capable of producing multi-spike train. In the second type, no such desired output spike train is specified beforehand and the algorithms choose the best time to spike for each pattern during training. Tempotron and the proposed algorithm are examples of this type which chooses $t_{max}$ (defined in Sec. \ref{sec.learning.algorithm}) as the time to spike. We have already compared the performance of our algorithm with Tempotron. We now choose ReSuMe as a representative of the first type of algorithms and compare its performance with the proposed learning rule for the two tasks described in Sec. \ref{task_descrip}. For ReSuMe, the neuron had to fire one spike at $t_+=350ms$ for $P^+$ patterns and at $t_-=450ms$ for $P^-$ patterns. The classification performance of ReSuMe (89.58\%(SD=1.24\%)) and 82.65\%(SD=2.22\%) accuracy for 500 and 1000 patterns) in Task I is worse compared to the proposed method. Similarly, for Task II, the proposed method performs much better than ReSuMe learning rule ($96.74\%(SD=0.8467\%)$ and $92.12\%(SD=1.1654\%)$ accuracy for 500 and 1000 patterns).

\subsection{Comparison to other classifiers using dendritic processing}
We have already compared the proposed algorithm with previous works employing NNLD in Section \ref{sec.intro}. Here we first compare our method with another dendritic algorithm proposed by Wu et al. in \cite{Wu&Mel2009}. Unlike \cite{Wu&Mel2009} which considered only mean rate encoded inputs, our learning rule can be applied to arbitrary spike trains. 

Second, we compare the proposed algorithm with another recently reported structural plasticity based algorithm named Dendritically Enhanced Readout (DER) \cite{Roy2013,LSM-DER}. In \cite{Roy2013,LSM-DER}, this structure has been used as the readout stage of Liquid State Machine. The primary difference of the proposed algorithm with DER is in the number of data points per pattern to be memorized for a particular task. If we consider there are $P^+$ and $P^-$ patterns of Class 1 and Class 2 respectively, then the number of data points to be memorized by the proposed algorithm are $P_{m\_NNLD}=T\times P^-+1$ while the number of data points to be memorized by DER are $P_{m\_DER}=T\times (P^++P^-)$ where $T$ is the number of time points per pattern. So, DER has to memorize almost two times more data points than the proposed algorithm for the same number of patterns thereby reducing its memory capacity.    

Third, we have also compared our performance with another recently proposed dendritic algorithm termed as Synaptic Kernel Inverse Method(SKIM) \cite{Tapson2013}. The difference mentioned above in the context of DER also applies to SKIM. Apart from that, SKIM also uses much more resources than the proposed NNLD which we will show next. We compare the two methods on Task I for $100$ patterns in which case our proposed method has 100\% accuracy. The neural network architecture used in \cite{Tapson2013} consists of $N$ presynaptic neurons which connect to an output spiking neuron, via synaptic connections to its $M$ dendritic branches. The weights of these synapses are random and fixed. These synapses along with a subsequent nonlinearity projects the input to a higher dimension thereby increasing separability. The dendritic branches sum the synaptic input currents, and the output from the dendritic branches are summed at the soma of the output neuron. The weights of the connection between dendritic branches and the soma are learnt by Moore pseudo inversion method \cite{Tapson2013}. Thus, for this network  $N\times M$ synaptic resources are required for connecting the $N$ presynaptic neurons to $M$ postsynaptic  dendritic branches and $M$ synapses are required for connecting the $M$ dendritic branches to the single output spiking neuron. The number of input spiking neurons are equal to the number of afferents in the input data which in our case is $500$.  Since we have used $500$ synapses for NNLD, so initially we keep the number of postsynaptic dendrites in SKIM as 1 ($M=1$) to match the number of resources used by us. But for $M=1$, SKIM fails miserably and provides only 50\% accuracy. When the number of dendritic branches are increased, SKIM provides better results and finally is able to provide 100\% accuracy when $M=360$. So to provide equivalent result as the proposed algorithm SKIM requires $361$ times more resources than our proposed NNLD.             

\section{CONCLUSION}\label{sec.Conclusion}
A morphological learning rule that can be used to find the optimal morphology of neurons with nonlinear dendrites (NNLD) and binary synapses is presented. The learning rule includes a novel threshold adaptation technique. To see the effectiveness of the proposed method, the NNLD trained with morphological rule is used to solve two classification tasks and one real world problem. The results depict that our proposed method with $1$ bit weights can achieve comparable performance to tempotron learning rule with 4-bit to 6-bit quantized weights. 

\section*{Acknowledgement}
The authors acknowledge helpful discussions with Prof. Nitish Thakor regarding tactile sensing.

\end{document}